\pdfoutput=1
\documentclass[11pt]{article}

\usepackage[preprint]{acl}

\usepackage{times}
\usepackage{latexsym}
\usepackage[T1]{fontenc}
\usepackage[scaled=0.9]{FiraSans}
\usepackage[utf8]{inputenc}
\usepackage{microtype}
\usepackage{inconsolata}
\usepackage{adjustbox}
\usepackage{ltablex}   % longtable + tabularx
\usepackage{array,graphicx}
\usepackage{arydshln}
\usepackage{amsmath}
\usepackage{tabularray}
\usepackage{color}
\usepackage[dvipsnames]{xcolor}
\usepackage{xspace}
\usepackage{cleveref}
\usepackage{enumitem}
\usepackage{threeparttable}
\usepackage{multirow}
\usepackage{tikz}
\usepackage{subcaption}
\usetikzlibrary{positioning}
\usepackage[edges]{forest}
\usepackage{booktabs}
\usepackage{rotating}
\usepackage{pifont}
\usepackage{wasysym}
\usetikzlibrary{arrows.meta, positioning}
\newcommand{\no}{\ding{53}}
\newcommand{\ok}{\ding{51}}

\newcommand{\ednative}{\CIRCLE}          % filled circle
\newcommand{\edanchored}{\LEFTcircle}        % half-filled circle  
\newcommand{\edadjacent}{\Circle}        % empty circle

\keepXColumns
\definecolor{dip}{HTML}{D62F2F}

\usepackage{tcolorbox}
\usepackage{fontawesome6}   % for the lightbulb icon
\tcbuselibrary{skins, breakable}

\definecolor{keynoteframe}{HTML}{E8C547}     % soft pastel yellow border
\definecolor{keynotebg}{HTML}{fffbe3}
\definecolor{keynotetitle}{HTML}{8B6914}     % warm brown text
\definecolor{keynotetitlebg}{HTML}{FFF1B8}   % pastel yellow title bg

\newtcolorbox{keynote}[1][]{
  enhanced,
  colback=keynotebg,
  colframe=keynoteframe,
  boxrule=1pt,
  arc=5pt,
  outer arc=5pt,
  left=8pt, right=8pt,
  top=6pt, bottom=6pt,
  fontupper=\normalsize,
  overlay={
    \node[anchor=south east, inner sep=4pt] 
      at (frame.south east) 
      {\textcolor{keynotetitle}{\Large\faLightbulb[light]}};
  },
  #1
}

\title{From Triage to Discharge: A Survey of NLP Tasks, Methods, and Open Challenges in the Emergency Department}

\author{
 \textbf{Dipankar Srirag}\quad\textbf{Aditya Joshi}\quad\textbf{Salil S. Kanhere}\quad\textbf{Padmanesan Narasimhan}
\\
University of New South Wales, Sydney, Australia
\\
\texttt{\{d.srirag, aditya.joshi, salil.kanhere, padmanesan\}@unsw.edu.au}
}

\begin{document}
\maketitle
\begin{abstract}

Emergency departments (EDs) operate under time pressure, generating multimodal data such as clinical conversations, triage notes, and discharge documents. Recent advances in natural language processing (NLP), particularly pretrained transformers and large language models, have created new opportunities to support language and time-intensive stages of emergency care. Yet existing surveys map clinical NLP across the broader hospital workflow or focus on specific tasks. This survey analyses 46 papers spanning the three phases of ED: triage, diagnosis, and disposition, covering tasks such as triage classification, clinical summarisation, automatic diagnosis, report generation, and discharge documentation. We examine modelling paradigms, evaluation practices, and emerging benchmarks and shared tasks. Across tasks, we identify common trends, including a shift from task-specific neural architectures to pretrained language models, growing interest in interactive clinical systems, and increasing attention to clinically grounded evaluation.  Finally, we detail open challenges such as limited generalisability, noisy clinical inputs, and workflow constraints that inform future ED-NLP research.
\end{abstract}

\section{Introduction}
\label{sec:intro}

The use of natural language processing (NLP) in medical domains has been widely studied~\cite{wu2020deep}. Modern NLP approaches, particularly following the development of Transformer-based large language models (LLMs), have enabled a range of clinical tasks, including summarisation~\cite{adams-etal-2021-whats}, dialogue modelling~\cite{xu2023medical}, information extraction~\cite{fornasiere-etal-2024-medical}, and clinical decision support~\cite{xu2024reasoning}. This survey focuses on a specific clinical context: emergency departments (EDs), where clinicians work under severe time pressure, often within a four-hour window from arrival to disposition~\cite{jones2010four, mason2012time}. ED care unfolds across three phases: \textit{triage and arrival}, \textit{diagnosis and assessment}, and \textit{patient disposition} (See Appendix~\ref{sec:app-phases}). These phases generate triage notes, chief complaints, patient speech, consultation notes, radiology and laboratory reports, and discharge summaries, comprising both structured and unstructured textual records that cover initial reports, intermediate clinical reasoning, and final decisions. Applying NLP techniques to ED data therefore has the potential to reduce the documentation burden, surface important clinical signals, and support consistent communication with the patient throughout the ED workflow. Appendix~\ref{sec:promise} provides a detailed motivation for NLP approaches in the ED.

\begin{table}[t!]
    % \centering
    % \small
    % \setlength{\tabcolsep}{4pt}
    \begin{adjustbox}{width=\linewidth, center}
        \begin{tabular}{lccccc}
            \toprule
            \textbf{Survey} & \textbf{Tri.} & \textbf{Dia.} & \textbf{Dis.} & \textbf{Task} & \textbf{Method} \\
            \midrule
            \citet{Klug2024}                    & \no & \ok & \ok & --  & --  \\
            \citet{valizadeh-parde-2022-ai}     & \no & \ok & \no & \ok & --  \\
            \citet{Stewart2023ApplicationsReview} & \ok & \no & \no & --  & \ok \\
            \citet{Tyler2024}                   & \ok & \no & \no & \no & \ok \\
            \citet{yao-yu-2026-llm} & -- & -- & -- & -- & \ok \\
            \hdashline
            \textbf{Ours}                       & \ok & \ok & \ok & \ok & \ok \\
            \bottomrule
        \end{tabular}
    \end{adjustbox}
    \caption{Comparison with prior surveys across ED phases (\textbf{Tri}age, \textbf{Dia}gnosis, \textbf{Dis}position) and analytical dimensions (NLP \textbf{Task} formulation, \textbf{Method} analysis). \ok\ = covered; \no\ = not covered; -- = partial.}
    \label{tab:survey-comparison}
\end{table}

% | Paper | Sources | Scope | Triage | Diag. | Disp. | NLP Task Formulation | Method analysis | 
% |----|----|----|----|----|----|----|----|
% |  | NLP Venues | NLP in Hospital Journey | N | Y | Y | P | P | 
% | [2] | NLP and AI Venues | Medical Dialogue Systems | N | Y | N | Y | P | 
% | [3] | MEDLINE, Embase, WoS, Scopus | NLP at Triage | Y | N | N | P | Y |
% | [4] | MEDLINE, Embase, WoS, Scopus | AI/ML at Triage | Y | N | N | N | Y |
% | **Ours** | **NLP/ML/AI + selected health** | **NLP across full ED workflow** | **Y** | **Y** | **Y** | **Y** | **Y** | 
% P = Partial; N = No; Y = Yes
% This table shows concretely that prior surveys either (i) cover the broader hospital pathway rather than the ED specifically [1], (ii) address a single task family such as medical dialogue [2], or (iii) analyse clinical outcomes without formulating NLP tasks or examining NLP evaluation practices across phases [3,4]. Our contribution is the NLP-task-and-evaluation view across the full ED workflow, which none of these provide. **We commit to adding this table to the final version.**

Existing surveys of Artificial Intelligence (AI) and NLP in emergency medicine, as described in Table~\ref{tab:survey-comparison}, provide useful overviews of use cases and clinical outcomes. However, they primarily draw on clinical settings and rarely examine methodological aspects relevant to NLP, such as task formulation, modelling assumptions, or evaluation protocols in a systematic way~\cite {Kirubarajan2020,Stewart2023ApplicationsReview,Tyler2024}. Surveys within the NLP community often focus on specific tasks or model architectures without assessing their evaluation practices or alignment with real-world workflow constraints~\cite{DiMartino2022,valizadeh-parde-2022-ai,wang-etal-2025-survey}. The two closest recent surveys share a workflow-level view but differ from ours in scope. \citet{Klug2024} catalogues clinical-NLP applications along the inpatient hospital pathway from admission to discharge, and \citet{yao-yu-2026-llm} analyse LLM-based multi-agent clinical systems around workflow observables and deployment-readiness framework. Neither is ED-specific, and neither examines how ED-phase constraints shape task formulation and evaluation. This fragmentation leaves open fundamental questions about how ED-oriented NLP tasks are formulated, how modelling paradigms are chosen, and whether evaluation practices adequately reflect operational realities in emergency care. 

To address this gap, we survey NLP research applied to ED workflows through a corpus of 46 papers\footnote{Search strategy and inclusion criteria in Appendix~\ref{sec:search-criteria}}. We analyse how these systems are deployed across ED phases and examine trends in dataset language, input modality, and modelling paradigms. We further synthesise task-level developments across triage classification, clinical interaction summarisation, automatic diagnosis, radiology report generation, disposition prediction, and discharge documentation. Finally, we discuss benchmarking resources and outline open challenges and future research directions for deploying NLP systems in ED settings.

\begin{figure*}[t!]
    \centering
    \includegraphics[width=0.9\linewidth]{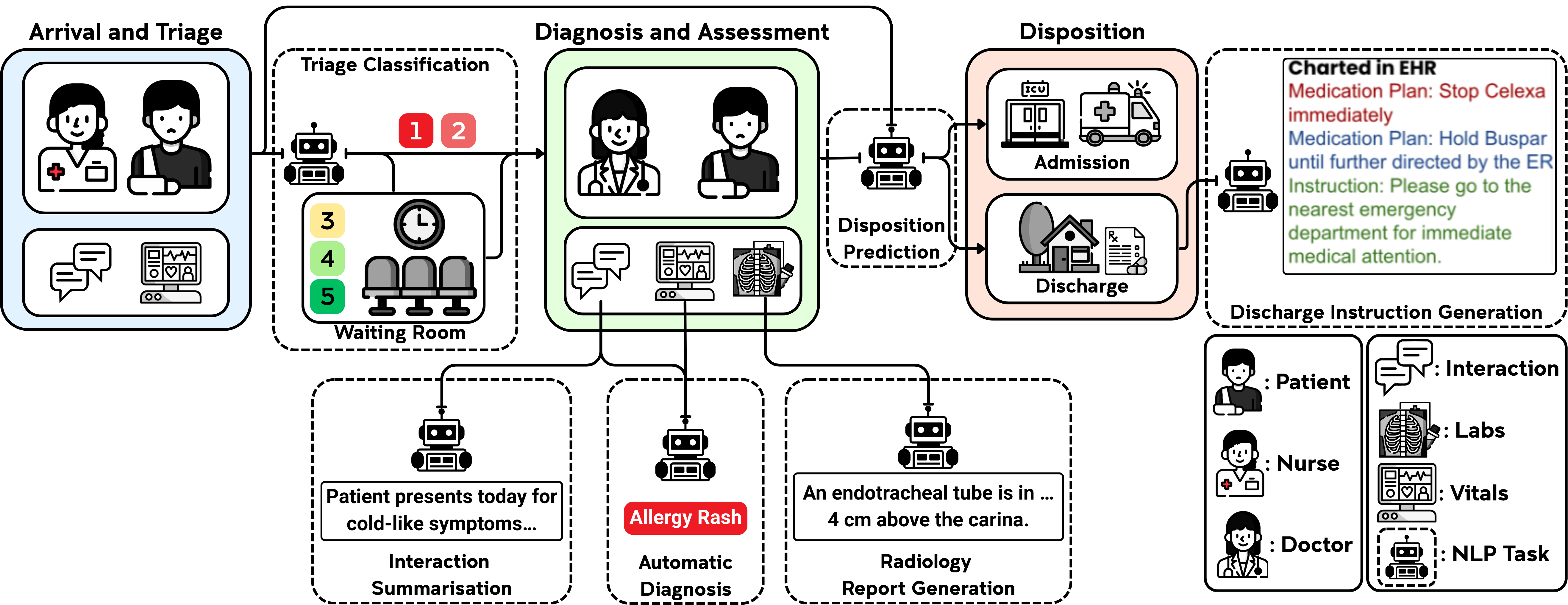}
    \caption{Overview of NLP tasks across the three phases of the ED workflow.}
    \label{fig:tasks}
\end{figure*}
\section{Research Trends}
\label{sec:trends}

Tables~\ref{tab:appendix_full_catalog_1} and~\ref{tab:appendix_full_catalog_2} in Appendix~\ref{sec:paper-list} provide a paper-level catalogue of all 46 included papers. We outline the trends across three dimensions: clinical settings, training paradigms, and evaluation paradigms.

\subsection{Clinical settings}

\paragraph{ED phase.}
Most work targets \textit{diagnosis and assessment} (35/46), where language data are relatively rich and naturally support tasks such as clinical summarisation, diagnostic reasoning, and report generation. By contrast, triage receives limited attention (5/46), despite being one of the most time-critical stages of ED care. Disposition tasks appear in 6 studies, including disposition prediction and discharge documentation. Research therefore concentrates on diagnostic reasoning and documentation rather than early triage decision support. Applying the ED-relatedness tags, the corpus splits into 10 ED-native, 17 ED-anchored, and 19 ED-adjacent papers. Papers corresponding to triage and disposition are almost entirely ED-native, whereas automatic diagnosis is dominated by ED-adjacent work (16/19). This distribution makes the inclusion boundary visible, as much of what is called ED-NLP in CS venues is methodologically transferable to ED but not evaluated there, which conditions the trend estimates that follow.

\paragraph{Language.}
English remains the dominant language (32/46) for NLP research in ED, followed by Chinese (13/46). Other languages, such as German and Spanish, appear in a small number of studies (3/46), and the source language is unspecified in 2/46. This imbalance largely reflects the availability of public clinical corpora, especially the MIMIC family of datasets~\cite{mimic-cxr,mimic-iv-ed,mimic-iv-note}, which have become standard resources for ED-facing clinical NLP. The limited representation of other languages highlights the scarcity of multilingual datasets and evaluation resources.

\subsection{Training paradigms}

\paragraph{Input modality.}
Text is the primary input modality (43/46), including triage notes, clinical conversations, physician documentation, and radiology reports. Structured electronic health record variables appear in 22/46 studies, while only 4/46 incorporate medical imaging directly. Despite the availability of heterogeneous multimodal information, most current work remains text-centred.

\paragraph{Training strategies.}
Supervised methods appear in 13/46 studies, agent-based systems in 10/46, transfer learning in 8/46, knowledge-grounded models in 6/46, pretraining in 4/46, reinforcement learning in 3/46, retrieval-augmented generation in 2/46, and prompt-based methods in only one paper. This distribution suggests a lack of a single dominant strategy beyond conventional supervised and transfer learning approaches. The scarcity of prompt-based methods in the corpus contrasts with their prevalence in shared task submissions (Section~\ref{sec:bench-tasks}). Agent-based and retrieval-augmented methods are concentrated in 2024 and 2025, reflecting the recent influence of LLM-based pipelines.

\subsection{Evaluation paradigms}

\paragraph{Study design.}
Most papers rely on retrospective records (24/46), simulation, or synthetic data (21/46). Only one study adopts a prospective design~\cite{Ip2024}, with data collected alongside live clinical care. As a result, most findings remain supported by offline evaluation rather than real-time clinical use.

\paragraph{Clinician evaluation.}
Only 11/46 papers supplement automated metrics with clinician or human judgement. The remainder rely entirely on automatic evaluation~\cite{lin-2004-rouge}, despite the limited alignment between such metrics and clinical utility~\cite{moramarco-etal-2022-human}. This gap is especially important in ED settings due to high cost of error.

\paragraph{Deployment.}
No paper in our corpus reports clinical deployment within an ED-specific workflow. The closest case, MTDiag~\cite{hou2023mtdiag}, is deployed in an online medical-consultation platform, which by our scope does not qualify as ED deployment. Real-world ED validation remains the most concrete open challenge for the field.
\section{NLP Tasks in the ED}
\label{sec:tasks}

NLP methods in healthcare support several functions that align closely with the three phases of the ED workflow. Figure~\ref{fig:tasks} summarises the main ED-focused tasks across the three ED phases. The following subsections describe each task, the main training strategies (Figure~\ref{fig:training_paradigms}), and the metrics used to evaluate them. Appendix~\ref{app:performance} provides a summary of reported performance across the reviewed tasks.

\paragraph{Common formulation.}
We formalise the ED NLP tasks under a common conditional prediction framework. Each task defines an input space $\mathcal{X}$ and an output space $\mathcal{Y}$, and models the conditional distribution parameterised by $\theta$: $\hat{y} = \arg\max_{y \in \mathcal{Y}} p_\theta(y \mid x)$. Tasks differ in what constitutes $x$ (the information available at the corresponding ED phase) and $y$ (the required output); each subsection below specifies both.

\subsection{Triage Classification}
\label{sec:triage}

Triage classification assigns an urgency or severity category~\cite{etek, esi} at \textit{arrival} and \textit{triage}. Under the common formulation, the output space is an ordered set of $K$ acuity levels: $\mathcal{Y} = \{1 \prec 2 \prec \cdots \prec K\}$ with $K=5$ under most triage scales (Appendix~\ref{sec:app-phases}), used for prioritisation and resource allocation~\cite{Stewart2023ApplicationsReview, Tyler2024}. The ordering $\prec$ reflects that acuity is ordinal (Level 1 is most urgent), though most work treats $\mathcal{Y}$ as a flat categorical space. Early work used topic modelling on triage notes to reveal symptom clusters and temporal patterns, suggesting that free text can support triage decisions~\cite{kocbek-etal-2014-exploring}. Existing work utilises several modelling paradigms, which can be grouped into transfer-learning approaches based on pretrained transformers, agent-based, and retrieval-augmented LLM systems.

\paragraph{Transfer learning approaches.} These formulate the task as supervised classification using contextual embeddings from pretrained transformers, with input $x = (c, v)$ combining the free-text chief complaint $c$ with vital signs $v$ serialised into textual form~\cite{maschhur-etal-2024-towards, Liu2025}. The resulting representation is then used for triage classification~\cite{maschhur-etal-2024-towards}, along with expert knowledge~\cite{Liu2025}. Overall, these approaches use pretrained encoders to jointly model short textual descriptions and structured triage variables.

\paragraph{Agent-based approaches.} Recent work explores LLMs acting as collaborative agents for triage decision-making, with input $x = (c, I)$ comprising the chief complaint $c$ and an instruction/prompt $I$. One line of work uses multiple LLM-based medical experts that iteratively discuss a case and converge on a triage decision through multi-round deliberation~\cite{lu-etal-2024-triageagent}. Another approach extends the input to $x_t = (c, v, h_t, I)$ at turn $t$, where $h_t$ is the dialogue history, modelling triage as a dynamic process that revises the assigned level as additional patient information becomes available~\cite{heal-www-2025}. This formulation explicitly models interacting with patients to gather additional information.

\paragraph{Retrieval-augmented approaches.} These augment LLMs with external medical knowledge, with input $x = (c, v, m, R, I)$ where $R$ is retrieved evidence and $m$ captures encounter metadata such as patient demographics. Methods serialise structured ED variables and clinical notes into prompts, then retrieve supporting evidence from sources such as PubMed\footnote{\url{https://pubmed.ncbi.nlm.nih.gov/}} to guide prediction~\cite{Gaber2025}. Experiments suggest that incorporating vital signs and retrieved references can improve performance across several LLM configurations. Comparative evaluations further indicate that general-purpose LLMs can match untrained ED practitioners in vignette-style settings, though they do not yet match professionally trained raters~\cite{Masanneck2024}.

\paragraph{Metrics.} Triage classification is usually evaluated with standard classification metrics such as accuracy, precision, recall, F1, and AUC~\cite{maschhur-etal-2024-towards, Liu2025, heal-www-2025, Gaber2025}. Some work also reports disagreement-based measures such as total discordance, defined as $1-\text{accuracy}$, to quantify mismatch with clinician-assigned triage levels~\cite{lu-etal-2024-triageagent}. However, most studies treat triage as a standard multi-class classification task and do not account for the ordinality of triage categories. Dataset properties such as long-tailed symptom distributions are also rarely addressed explicitly during model development~\cite{maschhur-etal-2024-towards}. 

\begin{keynote}
    Evaluations typically use post-hoc triage notes or EHR data that already contain information accumulated during the encounter rather than the limited inputs available at the moment of triage. This assumption inflates apparent accuracy and conflicts with how triage is performed in practice.
\end{keynote}

\definecolor{violetNode}{HTML}{E8DDF7}
\definecolor{indigoNode}{HTML}{E1E7FA}
\definecolor{blueNode}{HTML}{E3F1FF}
\definecolor{greenNode}{HTML}{E7F6E2}
\definecolor{yellowNode}{HTML}{FFF8D9}
\definecolor{orangeNode}{HTML}{FFEBD9}
\definecolor{redNode}{HTML}{FFE1DD}
\definecolor{magentaNode}{HTML}{F7DDF0}

\begin{figure*}[htbp]
\centering
\scriptsize

\tikzset{
  basic/.style={
    draw,
    rectangle,
    rounded corners=2pt,
    thin,
    inner sep=2pt
  },
  root/.style={
    basic,
    align=center,
    fill=gray!12,
    font=\bfseries,
    text width=1.55cm
  },
  catbox/.style={
    basic,
    align=center,
    font=\bfseries,
    text width=2.9cm,
    text depth=0pt,
    text height=1.2ex
  },
  leafbox/.style={
    basic,
    align=left,
    text width=9.2cm
  },
  midleaf/.style={
    basic,
    align=left,
    text width=5.9cm
  },
  shortleaf/.style={
    basic,
    align=left,
    text width=3.8cm
  },
  veryshortleaf/.style={
    basic,
    align=left,
    text width=1.8cm
  }
}

\begin{forest}
for tree={
  grow=east,
  parent anchor=east,
  child anchor=west,
  anchor=west,
  edge={-, thin},
  edge path={
    \noexpand\path[\forestoption{edge}]
    (!u.parent anchor) -- +(12pt,0) |- (.child anchor);
  },
  l sep=8mm,
  s sep=2mm
},
[{\textbf{Training}\\\textbf{strategies}},
  root,
  rotate=90,
  parent anchor=south,
  child anchor=west,
  anchor=center,
  [Pretraining, catbox, fill=violetNode,
    [{\citet{grambow-etal-2022-domain}; \citet{Chen2023}; \citet{yuan-etal-2024-continued}; \citet{wang-etal-2025-towards-adapting}}, leafbox, fill=violetNode!55]
  ]
  [Transfer Learning, catbox, fill=indigoNode,
    [{\citet{li-etal-2020-pharmmt}; \citet{lovelace-mortazavi-2020-learning}; \citet{krishna2021generating}; \citet{su2022extract}; \citet{chen-etal-2023-clinical}; \citet{singh-etal-2023-large}; \citet{maschhur-etal-2024-towards}; \citet{Liu2025}}, leafbox, fill=indigoNode!55]
  ]
  [Supervised, catbox, fill=blueNode,
    [{\citet{jeblee-etal-2019-extracting}; \citet{Fernandes2020}; \citet{joshi-etal-2020-dr}; \citet{rahman2020grace}; \citet{pmlr-v126-schloss20a}; \citet{li-etal-2022-diaformer}; \citet{hou2023mtdiag}; \citet{wang-etal-2023-coad}; \citet{xu2023medical}; \citet{Ip2024}; \citet{gatto-etal-2025-follow}; \citet{nicolson-etal-2025-impact}; \citet{qin-etal-2025-listening}}, leafbox, fill=blueNode!55]
  ]
  [Reinforcement Learning, catbox, fill=greenNode,
    [{\citet{wei-etal-2018-task}; \citet{fansi-etal-2022-towards-trustworthy}; \citet{sun-etal-2024-edcopilot}}, midleaf, fill=greenNode!55]
  ]
  [Knowledge-Grounded, catbox, fill=yellowNode,
    [{\citet{lin2021graph}; \citet{naseem2022incorporating}; \citet{eremeev-etal-2023-injecting}; \citet{xu2024reasoning}; \citet{jia2025medikal}}, leafbox, fill=yellowNode!60]
  ]
  [Retrieval Augmented, catbox, fill=orangeNode,
    [{\citet{Gaber2025}; \citet{xia2025mmedrag}}, shortleaf, fill=orangeNode!55]
  ]
  [Prompt-based, catbox, fill=redNode,
    [{\citet{nair2023generating}
    % ; \citet{liu-etal-2024-e}; \citet{socrates-etal-2024-yale}; \citet{wu-etal-2024-epfl}
    }, 
    veryshortleaf, fill=redNode!55]
  ]
  [Agent-based, catbox, fill=magentaNode,
    [{\citet{dou2023plugmed}; \citet{dou2024integrating}; \citet{he-etal-2024-bp4er}; \citet{lu-etal-2024-triageagent}; \citet{sun-etal-2024-edcopilot}; \citet{Chen2025}; \citet{heal-www-2025}; \citet{ju-lee-2025-prediction}; \citet{rose-etal-2025-meddxagent}; \citet{sun-etal-2025-enhancing-medical}}, leafbox, fill=magentaNode!55]
  ]
]
\end{forest}

\caption{Training strategies used in the included literature.}
\label{fig:training_paradigms}
\end{figure*}
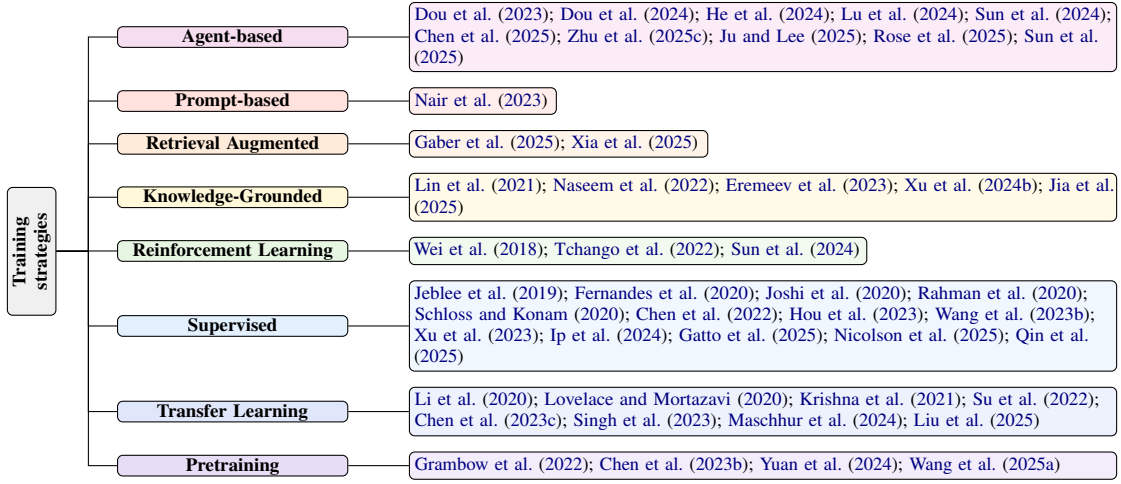
\subsection{Clinical Interaction Summarisation}
\label{sec:interaction-summarisation}

Clinical interaction summarisation converts multi-speaker doctor-patient conversations into structured clinical notes. Under the common formulation, the output space is a structured note partitioned into $S$ sections: $\mathcal{Y} = \{n = (n^{(1)}, \ldots, n^{(S)})\}$ most commonly in SOAP format (Subjective, Objective, Assessment, Plan; $S = 4$). Generation is typically autoregressive under a sequence-to-sequence model $p_\theta$. Existing work can be grouped into task-specific neural architectures, transfer-learning approaches based on pretrained transformers, domain-adaptive pretraining approaches, and prompt-based methods. These approaches input $x = d = (u_1, \ldots, u_T)$ a multi-turn dialogue where each utterance $u_t = (s_t, w_t)$ pairs a speaker label $s_t \in \{\text{clinician}, \text{patient}\}$ with a token sequence $w_t$.

\paragraph{Neural architectures.} Early work relied on task-specific neural architectures trained directly on annotated dialogue datasets. These methods often decomposed the task into intermediate steps such as utterance classification, clinical entity extraction, and SOAP section prediction~\cite{jeblee-etal-2019-extracting, pmlr-v126-schloss20a}. Other studies explored abstractive summarisation with pointer-generator networks~\cite{see-etal-2017-get} to better model negation and encourage copying of clinically salient content from the source dialogue~\cite{joshi-etal-2020-dr}. Pipeline-oriented systems were also developed to combine automatic speech recognition, clinical concept extraction, and rule-based post-processing for structured form filling from emergency medical service intake reports~\cite{rahman2020grace}. Overall, these methods were highly task-specific and often depended on multi-stage pipelines.

\paragraph{Transfer learning approaches.} These formulate summarisation as a transfer-learning problem, with the same input processed through pretrained encoders or encoder-decoder models to support both utterance-level note section assignment and section-conditioned summary generation~\cite{su2022extract, chen-etal-2023-clinical}. For example, \citet{krishna2021generating} propose a hybrid extractive-abstractive approach that identifies noteworthy utterances using a BERT-LSTM classifier and then generates section-conditioned summaries using a fine-tuned T5 model. Others adapt transformer models to conversational medical data before performing section classification~\cite{chen-etal-2023-clinical} or abstractive summarisation with pointer-generator networks~\cite{singh-etal-2023-large}. Compared with earlier systems, these methods rely less on handcrafted pipelines and more on transfer from LLMs.

\paragraph{Pretraining approaches.} A related line of work adapts pretrained models to the clinical dialogue domain before task-specific fine-tuning. This includes continued or domain-adaptive pretraining with denoising objectives on conversational or synthetic clinical corpora, followed by fine-tuning for note generation or note section summarisation~\cite{grambow-etal-2022-domain, Chen2023}. More recent work explores continual pretraining of LLMs on clinical corpora such as MIMIC-IV-Note~\cite{mimic-iv-note} before adapting them for summarisation tasks~\cite{yuan-etal-2024-continued}. \citet{wang-etal-2025-towards-adapting} further align summarisation quality with clinician expectations through reinforcement learning from human and AI feedback.

\paragraph{Prompt-based approaches.} Recent work explores prompt-based approaches using LLMs without task-specific training, extending the input to $x = (d, I)$ with an instruction/prompt $I$. \citet{nair2023generating} formulate the task as a multi-step prompting pipeline using GPT-3~\cite{brown2020languagemodelsfewshotlearners}, first extracting clinical entities and then generating structured summaries conditioned on these entities. Expert evaluation is used to assess the quality and coverage of the summary and clinical concepts.

\paragraph{Metrics.} Evaluation typically relies on automated metrics such as ROUGE~\cite{lin-2004-rouge} and BERTScore~\cite{Zhang2020BERTScore}, along with medical concept coverage via tools like QuickUMLS~\cite{grambow-etal-2022-domain, singh-etal-2023-large}. Some work also reports utterance-level section classification accuracy using metrics such as AUROC~\cite{jeblee-etal-2019-extracting, pmlr-v126-schloss20a}. \citet{savkov-etal-2022-consultation} show that grounding evaluations in pre-specified consultation checklists improves the correlation of these metrics with human judgements.

\begin{keynote}
    Approaches assume curated transcripts with clean turn boundaries, correct speaker labels, and complete utterances. Deployed ED summarisation would instead receive noisy ASR output with recognition and diarisation errors, informal patient phrasing, and frequent interruption.
\end{keynote}

\subsection{Automatic Diagnosis}
\label{sec:autodx}

Automatic diagnosis differs from the other tasks in being sequential, the model interleaves symptom inquiries with a terminal diagnosis over multiple turns. At turn $t$, given dialogue history $h_t = ((u_1, r_1), \ldots, (u_{t-1}, r_{t-1}))$ of prior inquiries $u$ and patient responses $r$, a policy $\pi_\theta$ selects either an inquiry action or a terminal diagnostic decision: $a_t = \pi_\theta(h_t)$, $a_t \in \mathcal{A}_{\text{inquire}} \cup \mathcal{D}$. $\mathcal{A}_{\text{inquire}}$ is the space of symptom inquiries and $\mathcal{D}$ is the set of possible diagnoses. In the ED, this aligns with \textit{diagnosis and assessment}, where structured history-taking and timely rule-in or rule-out decisions are critical. We focus on approaches trained and evaluated on derived data (dialogues) rather than on single-turn settings with fully observed information~\cite{medQA}. Existing work can be grouped into reinforcement learning, supervised, knowledge-grounded, and agent-based approaches. Surveys on medical agents provide complementary categorisations of this literature~\cite{shi-etal-2024-medical, wang-etal-2025-survey}. Approaches differ in how $\pi_\theta$ is parameterised and trained.

\paragraph{Reinforcement learning approaches.} These formulate diagnosis as sequential decision-making and learn $\pi_\theta$ through interaction rewards. \citet{wei-etal-2018-task} propose a task-oriented dialogue system that uses a Deep Q-Network to decide whether to inquire about additional symptoms or produce a diagnosis. Later work extends this paradigm with more structured reward design. For example, \citet{fansi-etal-2022-towards-trustworthy} introduce a dual-branch architecture combining an evidence acquisition policy with a disease classifier and reward functions that encourage exploration of differential diagnoses and prioritisation of severe conditions. Despite these advances, such approaches often require carefully engineered reward functions and can suffer from unstable training and limited data efficiency.

\paragraph{Supervised approaches.} These reformulate diagnosis as supervised prediction, learning $\pi_\theta$ from labelled interactions using transformer-based architectures~\cite{li-etal-2022-diaformer, wang-etal-2023-coad, hou2023mtdiag}. Some treat diagnosis as symptom-sequence generation~\cite{li-etal-2022-diaformer}, others use collaborative or multi-task learning to align symptom checking with disease prediction~\cite{wang-etal-2023-coad, hou2023mtdiag}. More recent work models dialogue structure, follow-up questions, and patient responses to improve symptom tracking and downstream reasoning~\cite{xu2023medical, gatto-etal-2025-follow, qin-etal-2025-listening}. However, supervised approaches often assume symptoms in $h_t$ are available in structured form, requiring extraction components to recover them from conversation.

\paragraph{Knowledge-grounded approaches.} These augment $h_t$ with structured medical knowledge to improve diagnostic reasoning. Prior work injects medical knowledge graphs into transformer models through knowledge-enriched utterance representations~\cite{naseem2022incorporating}, while graph-based methods model symptom-disease relationships with evolving knowledge graphs to guide dialogue generation and entity prediction~\cite{lin2021graph}. More recent approaches integrate knowledge graphs with LLMs. For example, \citet{jia2025medikal} propose medIKAL, which combines LLM predictions with knowledge graph retrieval and path-based re-ranking to identify candidate diseases. Other work focuses on improving how LLM reasoning better reflects clinical diagnostic processes by explicitly modelling intermediate reasoning steps during dialogue~\cite{xu2024reasoning}.

\paragraph{Agent-based approaches.} These implement $\pi_\theta$ as LLM agents, often decomposing diagnosis into specialised functions such as symptom acquisition, knowledge retrieval, and differential ranking~\cite{rose-etal-2025-meddxagent, ju-lee-2025-prediction, dou2023plugmed}. Some approaches inject structured diagnostic protocols via preference learning~\cite{dou2024integrating} or coordinate multi-agent multidisciplinary consultations~\cite{Chen2025}. Prompting-based variants expose intermediate reasoning~\cite{he-etal-2024-bp4er} or dynamic prompt adaptation with retrieved knowledge~\cite{sun-etal-2025-enhancing-medical}.

\paragraph{Metrics.}
Evaluation of automatic diagnosis systems typically measures diagnostic accuracy and symptom acquisition efficiency. Common metrics include disease prediction accuracy, recall of implicit symptoms, and the number of conversational turns required to reach a diagnosis~\cite{wei-etal-2018-task,fansi-etal-2022-towards-trustworthy}. Some work also evaluates ranking-based metrics such as Recall@K for differential diagnosis generation~\cite{li-etal-2022-diaformer,wang-etal-2023-coad}. Reinforcement learning approaches additionally report success rate and cumulative reward to measure efficiency. However, few studies evaluate agreement with clinician diagnoses or assess performance in real clinical workflows. Most rely on simulated patients or controlled retrospective settings, raising questions about reliability and privacy when deployed against real ED encounters. Recent work documents systematic biases in LLM-based disease prediction across gender and age subgroups~\cite{zhao-etal-2024-llms} and broader hallucination risks in clinical LLMs~\cite{zhu-etal-2025-trust}, both of which compound in ED settings.

\begin{keynote}
Sequential formulations optimise for symptom acquisition efficiency, yet real ED diagnosis is bounded by clinician time rather than inquiry count. The turn-count rewards produces models that minimise interaction, and leads to a mismatch neither the training objective nor the evaluation captures.
\end{keynote}

\subsection{Radiology Report Generation}
\label{sec:reportgen}

Prior surveys reviewed medical report generation more broadly, including consultations and discharge summaries and their evaluation~\cite{zhou-etal-2023-survey}. In contrast, we focus on \textit{radiology report generation} (RRG) for chest X-rays, emphasising methods and metrics that matter in the ED context. The output space is a structured report with two sections: $\mathcal{Y} = \{r = (r_{\text{find}}, r_{\text{impr}})\}$ where $r_{\text{find}}$ is the free-text \textit{Findings} section describing observed abnormalities and $r_{\text{impr}}$ is the \textit{Impression} section summarising clinical conclusions. Most models are trained and evaluated on large CXR corpora such as MIMIC-CXR~\cite{mimic-cxr}. Because many of these examinations occur during ED stays, the task is directly relevant to ED workflows. 

Early work adapted image-captioning architectures to radiology with input $x = G$, a chest radiograph. \citet{lovelace-mortazavi-2020-learning} extract spatial features with a pretrained DenseNet-121 encoder and add an auxiliary objective that enforces agreement between clinical observations extracted from generated and reference reports. More recent work extends the input to $x = (G, \zeta)$ with clinical context $\zeta$ (triage vitals, medications, indication, prior 
imaging). CXRMate-ED~\cite{nicolson-etal-2025-impact} combines chest X-rays with triage vitals and medications, embedding heterogeneous inputs to prompt a Llama-based decoder with a UniFormer image encoder, trained via supervised learning and self-critical sequence training. Retrieval-augmented approaches further extend the input to $x = (G, \zeta, R)$ where $R$ is retrieved evidence. MMed-RAG~\cite{xia2025mmedrag} filters retrieved evidence before generation and applies preference-based fine-tuning to align outputs with reference reports. Together, these methods reflect a shift toward context-aware and retrieval-grounded generation to reduce clinically harmful hallucinations.

\paragraph{Metrics.} Evaluation of radiology report generation typically relies on automated text generation metrics such as ROUGE, BLEU~\cite{papineni-etal-2002-bleu}, and CIDEr~\cite{vedantam2015cider}, along with precision and recall for the extracted clinical concepts. Some recent work also reports diagnostic classification metrics when assessing the clinical correctness of generated reports~\cite{xia2025mmedrag}. \citet{zhao-etal-2024-ratescore} introduce an entity-aware metric that decomposes reports into typed medical entities and compares them via embedding similarity, showing closer alignment with expert preference. The risks of persistent hallucinations mean that factual consistency checks and dedicated error detection remain essential for ED deployment~\cite{min-etal-2022-rred, rusak-etal-2023-catching}. 

\begin{keynote}
    Current methods populate $\zeta$ with structured triage variables (vitals, medications), but radiologists reading ED chest X-rays also draw on the referring clinician's suspected diagnosis, prior imaging, and the specific clinical question being asked. Reports generated without this indication risk being clinically fluent but diagnostically off-target for the question that prompted the imaging.
\end{keynote}

\subsection{Tasks in Patient Disposition}
\label{sec:disposition}

Disposition in the ED spans two language-relevant problems. The first is \textit{disposition prediction}, which forecasts whether a patient will be admitted or discharged and the likely destination, such as a ward or ICU. The second is \textit{discharge instructions}, which translate care plans into patient-facing guidance. Both problems draw on resources available at triage time, such as early triage narratives, chief complaint text, brief histories, clinician notes, vitals and demographics. We structure this section by task as the literature is relatively small.

\paragraph{Disposition Prediction.} Disposition prediction aims to identify patients at risk of hospitalisation or clinical deterioration early in the ED encounter. The output is a categorical disposition class 
$y \in \mathcal{D} = \{\textit{discharge}, \textit{ward}, \textit{ICU}\}$. Early work uses $x = (c, v, m)$ combining routine triage variables with free-text chief complaints, showing that textual features complement structured variables for outcomes such as ICU admission and related critical events~\cite{Fernandes2020}. Multimodal variants extend the input to $x = (c, v, m, V)$ with lightweight visual signals $V$; for example, short triage videos capturing clinical gestalt improve hospitalisation prediction when fused with standard triage variables~\cite{Ip2024}. More recent work uses $x = (c, v, m, I)$ with LLM-based approaches that linearise heterogeneous patient data into a single textual prompt: ED-Copilot processes such input via BioGPT~\cite{Luo_2022} to predict critical outcomes and recommend informative laboratory tests~\cite{sun-etal-2024-edcopilot}, further using reinforcement learning to select laboratory groups that reduce diagnostic delay.

\paragraph{Discharge Instructions Generation.} At the point of disposition, language technologies simplify and standardise care plans into patient-facing instructions. The output is a patient-facing instruction sequence $y = \iota$ derived from clinical inputs. PharmMT~\cite{li-etal-2020-pharmmt} takes $x = p$, a prescription direction, and formulates simplification as neural machine translation from clinical to lay phrasing, reporting substantial gains on automatic metrics together with pharmacist-judged safety and usability at scale. For broader discharge instructions, \citet{eremeev-etal-2023-injecting} take $x = (d, K)$ where $d$ is a doctor-patient dialogue and $K$ is injected domain knowledge; this amplifies the use of rare clinical tokens and improves factuality and coherence compared with instruction-tuned baselines.

\paragraph{Metrics.} For disposition prediction, common metrics include AUROC, AUPRC, and calibration measures, often with subgroup analyses for higher- and lower-acuity patients~\cite{Fernandes2020, Ip2024, sun-etal-2024-edcopilot}. For discharge instructions, these are typically complemented by readability assessments and pharmacist or clinician judgements of utility and safety~\cite{li-etal-2020-pharmmt, liu2022retrieve, eremeev-etal-2023-injecting}.

\begin{keynote}
    Disposition prediction models optimise for aggregate AUROC, but the clinical cost of errors is asymmetric. Under-triage of an ICU-bound patient carries far greater consequences than over-triage of a ward-bound one. Discharge instruction generation, by contrast, produces the same output $\iota$ regardless of patient literacy, language, or cognitive state. Hence, across both tasks, evaluation should remain sensitive to calibration and subgroup performance in the heterogeneous ED population.
\end{keynote}

\section{Benchmarks and Shared Tasks}
\label{sec:bench-tasks}
Early benchmarks for clinical language models relied mainly on exam-style question answering datasets that measured factual recall rather than real clinical decision-making and communication~\cite{Jin2021,jin-etal-2019-pubmedqa}. Recent benchmarks explore open-ended tasks that better reflect clinical workflows, including summarisation, diagnosis, and disposition prediction. ClinicBench~\cite{liu-etal-2024-large} represents this shift by aggregating datasets for clinical language understanding, reasoning, and generation. SOTA LLMs report worse performance on open-ended tasks than on exam-style benchmarks. Benchmarks have also expanded beyond text-only evaluation toward multimodal and process-centric settings. MC-BEC~\cite{chen2023multimodal} integrates structured intake data, radiology reports, vital signs, and physiological waveforms across $>$100K ED visits for time-sensitive prediction tasks such as decompensation and disposition. Other resources target safety and longitudinal reasoning: MEDEC~\cite{ben-abacha-etal-2025-medec} evaluates detection and correction of clinically significant documentation errors, ACI-Bench~\cite{Yim2023} evaluates dialogue-to-note generation from full doctor-patient conversations. Finally, MedJourney~\cite{wu-etal-2024-medjourney}, MediQ~\cite{li2024mediq}, and DDxGym~\cite{winter-etal-2024-ddxgym} assess patient-journey and interactive diagnostic reasoning.

Shared tasks provide controlled comparisons under common datasets and protocols. In clinical interaction summarisation, the MEDIQA-Chat 2023 shared task introduced section-wise and full-note generation from doctor-patient conversations \citep{ben-abacha-etal-2023-overview}. Systems ranged from fine-tuned transformer models with dialogue-aware or retrieval-based components to in-context learning approaches based on LLMs. Results suggest that in-context learning is competitive with fully fine-tuned systems for sectioned note generation \citep{giorgi-etal-2023-wanglab,tang-etal-2023-gersteinlab}, while the released datasets have also enabled analysis of faithfulness and error patterns in dialogue-to-note generation \citep{ben-abacha-etal-2023-empirical}. 

A similar pattern appears in discharge documentation. The BioNLP 2024 Clinical Text Generation challenge introduced the \textit{Discharge Me!} task for generating brief hospital course and discharge instructions from EHR data \citep{xu-etal-2024-overview}. Submitted systems explored prompt-based LLMs, encoder-decoder fine-tuning, and hybrid pipelines combining structured extraction with controlled generation \citep{damm-etal-2024-wispermed,socrates-etal-2024-yale,liu-etal-2024-e,wu-etal-2024-epfl}. Dynamic input filtering was reported to improve both automatic scores and clinician-rated completeness and correctness \citep{wu-etal-2024-epfl}. Automatic metrics capture only part of the overall quality, making clinician evaluation necessary to assess correctness, completeness, and readability.

\section{Conclusion and Future Directions}

This survey reviewed recent advances in NLP for the ED workflow. We covered representative modelling approaches and evaluation metrics for key language-intensive tasks in ED settings: triage classification, clinical interaction summarisation, automatic diagnosis, radiology report generation, disposition prediction, and discharge instruction generation. We also discussed emerging benchmarks and shared tasks that move evaluation beyond exam-style question answering toward more realistic clinical workflows. Across these tasks, several common trends emerge: (i) recent work has shifted from task-specific neural architectures toward pretrained transformer models and LLMs; (ii) evaluation settings are increasingly open-ended, multimodal, and process-oriented, reflecting the complexity of real-world clinical workflows; (iii) many systems are no longer framed as isolated prediction models, but as interactive tools for summarisation, reasoning, and decision support. Despite this progress, our review surfaces several gaps that we believe should shape ED-NLP research.

\paragraph{Data limitations and generalisability.}
Many studies rely on single-institution datasets~\cite{mimic-iv-ed} or online health communities~\cite{dxy-aaai-2019}, raising concerns about ecological validity. Documentation style, patient population, case mix, and local workflow vary substantially across institutions, so broader multi-institution evaluation and purpose-built ED corpora outside the MIMIC family are needed for generalisability. Simulation frameworks such as TriageSim~\cite{srirag2026triagesimconversationalemergencytriage} generate persona-conditioned triage dialogue and aligned audio from structured EHR seed data. This offers one route to working around regulatory constraints on direct patient recording.

\paragraph{Robustness to real-world clinical data.}
Many systems assume clean and structured inputs: summarisation models rely on curated transcripts, while triage and diagnosis models assume complete symptom representations. ED data is neither clean nor complete: transcripts contain ASR and diarisation errors~\cite{Goss2016}, triage decisions are made under time pressure~\cite{hitchcock-2014-triage}, and patient descriptions are often informal or across language barriers~\cite{Meeuwesen2010}. Out-of-domain analysis in our corpus~\cite{chen-hirschberg-2024-exploring} shows substantial degradation from missing information and hallucination rather than format mismatch.

% \paragraph{Multimodal clinical reasoning.}
% Most current approaches rely primarily on text, even though clinical decision-making depends on multiple information sources, including medical images, laboratory results, physiological signals, and structured EHR data. This limitation is especially important for tasks such as automatic diagnosis, disposition prediction, and radiology report generation, where clinically relevant evidence is distributed across modalities. Developing models that can integrate multimodal and heterogeneous clinical signals, such as vital signs and conversational inputs, remains an important direction for building more realistic ED decision-support systems.

\paragraph{Evaluation and clinical validation.}
Current evaluation practices are heavily dependent on automated metrics such as accuracy, ROUGE, BLEU, or concept coverage scores (Section~\ref{sec:trends}). While these enable benchmarking, they correlate poorly with clinical usefulness and safety~\cite{moramarco-etal-2022-human, savkov-etal-2022-consultation}. Three concrete shortcomings recur across the corpus. First, almost all results are retrospective or simulated, which limits what can be claimed about real-time clinical use. Second, triage labels are ordinal but are almost always evaluated as flat multi-class classification (Section~\ref{sec:triage}), so reported accuracies do not distinguish small misclassifications from clinically dangerous ones. Future works should instead utilise weighted-$\kappa$ or rank-correlation metrics. Third, clinician evaluation is reported in only 11 of 46 papers and rarely uses pre-specified protocols. Checklist-based protocols~\cite{savkov-etal-2022-consultation} measurably improve both inter-rater agreement and the alignment of automatic metrics with human judgement, and should become a default for ED-facing generation tasks. Evaluations are typically performed using data assembled post-encounter rather than the limited inputs available in real time. Recent benchmarks such as EHR2Dial-Triage~\cite{zhu2026elicitedehrgroundedlongitudinalinteractive} address this by linking each patient disclosure to its supporting EHR event and the dialogue turn at which it becomes available.

\paragraph{Deployment, privacy and workflow integration.} 
No paper in our corpus reports clinical deployment within an ED-specific workflow (Section~\ref{sec:trends}), making deployment the most concrete open challenge. Many approaches rely on API-based models, conflicting with privacy regulations. Even locally deployable systems must integrate with EHR infrastructure while remaining transparent and accountable. Concerns around privacy~\cite{11142715}, bias~\cite{zhao-etal-2024-llms}, hallucination~\cite{zhu-etal-2025-trust}, and explainability~\cite{Amin2026} remain major barriers. Closing this gap will require closer collaboration among NLP researchers, clinicians, and healthcare institutions, alongside more realistic benchmarks and clinically grounded evaluation, so that the next generation of ED-NLP systems can move from offline benchmarks to real-time clinical impact.

\section*{Limitations}
This survey focuses on a defined subset of clinical NLP problems in the ED, specifically tasks aligned with triage, diagnosis and assessment, and disposition. As a result, it does not aim to cover the broader clinical NLP landscape exhaustively, and adjacent problems such as outpatient triage, inpatient documentation, telemedicine consultations, and post-discharge follow-up fall outside our scope (Appendix~\ref{sec:search-criteria}).

The corpus reflects the language and geographic skew of publicly available ED corpora. Our findings about methodological trends and evaluation practices therefore apply most directly to English- and Chinese-language ED-NLP research, and conclusions about generalisability to other languages and care settings should be drawn cautiously. Similarly, our discussion of radiology report generation is restricted to chest X-rays because that is where MIMIC-CXR concentrates the available ED-adjacent data, and our coverage of other imaging modalities is correspondingly limited.

Finally, this survey synthesises studies that differ substantially in task formulation, dataset design, and evaluation methodology. We report reproduced or self-reported numbers from the original papers where available (Appendix~\ref{app:performance}), but we deliberately avoid head-to-head benchmark rankings across heterogeneous setups. As a result, this work is better suited to identifying broad methodological trends and open challenges than to making strong comparative claims about real-world clinical performance.

\section*{Ethical Considerations}
The methods discussed in this survey should be treated as supplementary tools for clinicians rather than replacements for clinical expertise. In emergency care, incorrect, incomplete, or poorly calibrated model outputs could contribute to unsafe decisions, and the fluent surface form of LLM-generated text may encourage over-reliance on system recommendations even when the underlying reasoning is unsound. Patient-facing applications, such as discharge instruction generation, require particular caution because errors may directly influence patient understanding and adherence, with disproportionate impact on patients with lower health literacy or limited English proficiency.

Additional risks include privacy concerns arising from API-based deployments of clinical NLP systems, demographic bias in disease prediction and triage models, and limited generalisability across institutions, languages, and workflows. The ED setting compounds these risks because decisions are time-pressured, the patient population is heterogeneous and often vulnerable, and the cost of an under-triage or missed-diagnosis error is high. These concerns reinforce the need for clinician oversight, prospective validation in the target deployment setting, subgroup-aware evaluation, and stronger institutional scrutiny before real-world use.

\section*{Acknowledgements}
Dipankar Srirag is supported by NHMRC Ideas Grant
RG241647, awarded to Padmanesan Narasimhan and Aditya
Joshi in 2025. A privacy-preserving AI tool was used to assist with revising portions of the text (sentence structures). All content was manually revised and verified by the authors.

\bibliography{custom}
\appendix
% \begin{figure*}[t!]
%     \centering
%     \includegraphics[width=\linewidth]{figures/phases.png}
%     \caption{Overview of the main phases of the emergency department workflow.}
%     \label{fig:phases}
% \end{figure*}
\section{Emergency Department Phases}\label{sec:app-phases}

To contextualise the NLP tasks reviewed later, this section outlines the three main phases of the ED workflow: triage and arrival, diagnosis and assessment, and patient disposition. These phases structure how information is collected, documented, and acted upon during an ED encounter, and they motivate the task groupings used in the remainder of the survey.
\subsection*{Triage and Arrival}

As shown in Figure~\ref{fig:tasks}, the ED workflow begins with patient arrival, whether by self-presentation or emergency services transport. At this stage, a triage nurse rapidly evaluates the patient's condition and assigns an appropriate triage category. While some systems also employ a general practitioner, triage in most EDs is performed exclusively by nurses, who follow structured protocols rather than diagnostic reasoning. This distinction matters for NLP, since tools designed for triage must align with protocol-driven decision rules rather than physician-style assessment. Several triage scales are used internationally, including the Emergency Severity Index (ESI) in the United States~\cite{ESPEJO202557}, the Australasian Triage Scale (ATS) in Australia~\cite{chamberlain2015identification}, the Manchester Triage Scale in Europe~\cite{cooke1999does}, and the CETEC in China~\cite{Liu2025}. While these scales share a five-level ordinal structure (with Level 1 denoting the most urgent cases), their underlying criteria differ. For instance, the ESI assigns triage levels based on the anticipated resource requirements, whereas the ATS focuses on the maximum acceptable waiting time before physician assessment. Beyond triage, clinicians generate brief intake notes summarising the chief complaint, relevant history, vital signs, and the assigned category, which form the basis for subsequent decision-making.

\subsection*{Diagnosis and Assessment}

Following triage, patients progress to the diagnosis and assessment phase, which forms the core of clinical decision-making in the ED. Physicians combine information from the initial hand-off notes with elicited symptoms, examination findings, and diagnostic investigations such as laboratory tests or imaging. Documentation at this stage often follows the SOAP structure, consisting of \textit{Subjective} (patient-reported complaints), \textit{Objective} (observed signs and results), \textit{Assessment} (provisional or confirmed diagnosis), and \textit{Plan} (treatment and disposition). Depending on clinical presentation and local workflow, patients may be observed for a short stay before a final disposition decision is made.

\subsection*{Patient Disposition}

The final phase of an ED encounter involves disposition, where the clinician formalises the next stage of care. Patients may be discharged with prescriptions and follow-up instructions or admitted to an inpatient ward or intensive care unit. The discharge summary consolidates information about presenting complaints, diagnostic findings, treatment plans, and recommendations for ongoing management. Beyond serving as documentation, these summaries are critical for ensuring continuity of care across providers.

\section{The Promise of NLP in the ED}
\label{sec:promise}

Over the years, language technologies have progressed from sparse vector models to large-scale neural sequence models. Early approaches relied on bag-of-words representations, such as TF-IDF and n-gram language models, for tasks such as document retrieval, topic detection, and basic text classification. These were followed by linear and probabilistic models, including logistic regression, maximum entropy models, and conditional random fields, which improved performance on sequence labeling tasks such as part-of-speech tagging and named entity recognition. A major shift came with distributed word representations, in which neural language models and word embeddings replaced sparse features with dense vectors that encode semantic similarity. These representations enabled neural architectures, particularly recurrent networks such as LSTMs and GRUs, to become the dominant paradigm across a wide range of applications, including machine translation, dialogue state tracking, and extractive summarisation. Sequence-to-sequence models with attention further unified many tasks under a common encoder-decoder formulation, where the same architecture could be instantiated for translation, abstractive summarisation, and question answering by changing the training data and objective.

Transformer architectures and large-scale pretraining have since redefined the state of the art for both natural language understanding (NLU) and natural language generation (NLG). Encoder-only models trained with masked language modelling objectives~\cite{devlin-etal-2019-bert} are now routinely fine-tuned on multi-task benchmarks such as GLUE~\cite{wang-etal-2018-glue} and SuperGLUE~\cite{10.5555/3454287.3454581} for sentence classification, textual entailment, and natural language inference. Decoder-only~\cite{touvron2023llama2openfoundation} and encoder-decoder models~\cite{10.5555/3455716.3455856} trained with causal or span-corruption objectives underpin contemporary systems for open-domain question answering~\cite{zhang-etal-2023-survey-efficient}, dialogue summarisation~\cite{zhu-etal-2025-factual}, and machine translation~\cite{lyu-etal-2024-paradigm}. These models benefit from training on orders of magnitude more text than earlier systems and from parameter counts that support rich internal representations of context, discourse structure, and world knowledge. LLMs extend this trend by scaling data, model size, and training objectives to the point where a single model can act as a general-purpose language engine. When prompted with natural language instructions and a few examples, LLMs can perform new tasks with little or no task-specific fine-tuning, a capability often described as in-context learning~\cite{brown2020languagemodelsfewshotlearners}. They show strong performance on broad multitask benchmarks that cover knowledge-intensive and reasoning-heavy problems~\cite{she-etal-2023-scone}, as well as specialised benchmarks that evaluate code-generation and software-engineering capabilities, such as program synthesis and issue resolution~\cite{jimenez2024swebench}. In addition, tool-augmented variants integrate retrieval~\cite{10.5555/3495724.3496517}, structured reasoning~\cite{dong2026structured}, and external application programming interfaces~\cite{Plaat2025, polyakov-etal-2025-toolreflection}, enabling them to combine symbolic computation with free-text generation~\cite{freitas-etal-2025-neuro}.

Building on these advances, a growing body of work has focused on clinical NLP, where models are trained or adapted on electronic health records, clinical notes, radiology reports, and biomedical question answering datasets. Domain-specific pretraining and continued pretraining on de-identified clinical corpora have led to encoders and generative models that better capture medical terminology, abbreviations, and documentation conventions than general-purpose LMs. Evaluations on standard clinical benchmarks for concept extraction, document classification, natural language inference, and medical question answering consistently show gains when using such specialised models, particularly for tasks that require detailed knowledge of clinical workflows and disease-specific language~\cite{10.1145/3611651, liévin2023largelanguagemodelsreason, Singhal2023}. From a systems perspective, these models reuse the same architectural components as general-purpose LLMs, i.e., large-scale pretraining, task-specific fine-tuning or instruction tuning, retrieval augmentation, and safety layers, but with training data, ontologies, and evaluation criteria tailored to healthcare~\cite{10.1145/3611651}. 

Research on NLP applied to ED clinical data has predominantly focused on predicting outcomes from unstructured free-text triage notes, such as assignment of triage score, need for admission, or critical illness. A recent scoping review of NLP at emergency department triage found that, although NLP models can achieve high predictive accuracy for clinically relevant outcomes and combining free-text with structured data often improves results, the majority of studies exhibited high risk of bias and were retrospective in design. Moreover, only a single study reported actual deployment of an NLP model into clinical practice, highlighting a gap between retrospective performance and real-world impact~\cite{Stewart2023ApplicationsReview, wu-etal-2024-clinical-dataset-survey}. This pattern illustrates both the \textit{promise} and \textit{limitation} of current ED NLP: models can exploit narrative clinical content effectively, yet their validation and integration into clinical workflows remain limited.

Despite widespread adoption of language technologies in other healthcare domains~\cite{valizadeh-parde-2022-ai}, their direct application to the ED remains limited. Short, data-sparse encounters and the need for rapid documentation constrain methods that depend on longitudinal histories or rich contextual data. Privacy, safety, and workflow integration requirements also make it harder to reuse generic language models without adaptation. These conditions underscore the need for LTs that can operate under time pressure, efficiently integrate multimodal information, and support timely communication between clinicians. Potential applications include triage note classification, automated summarisation of clinician notes, dialogue systems for patient intake, and decision-support tools tailored for real-time environments.
\begin{figure}[t!]
    \centering
    \includegraphics[width=0.85\linewidth]{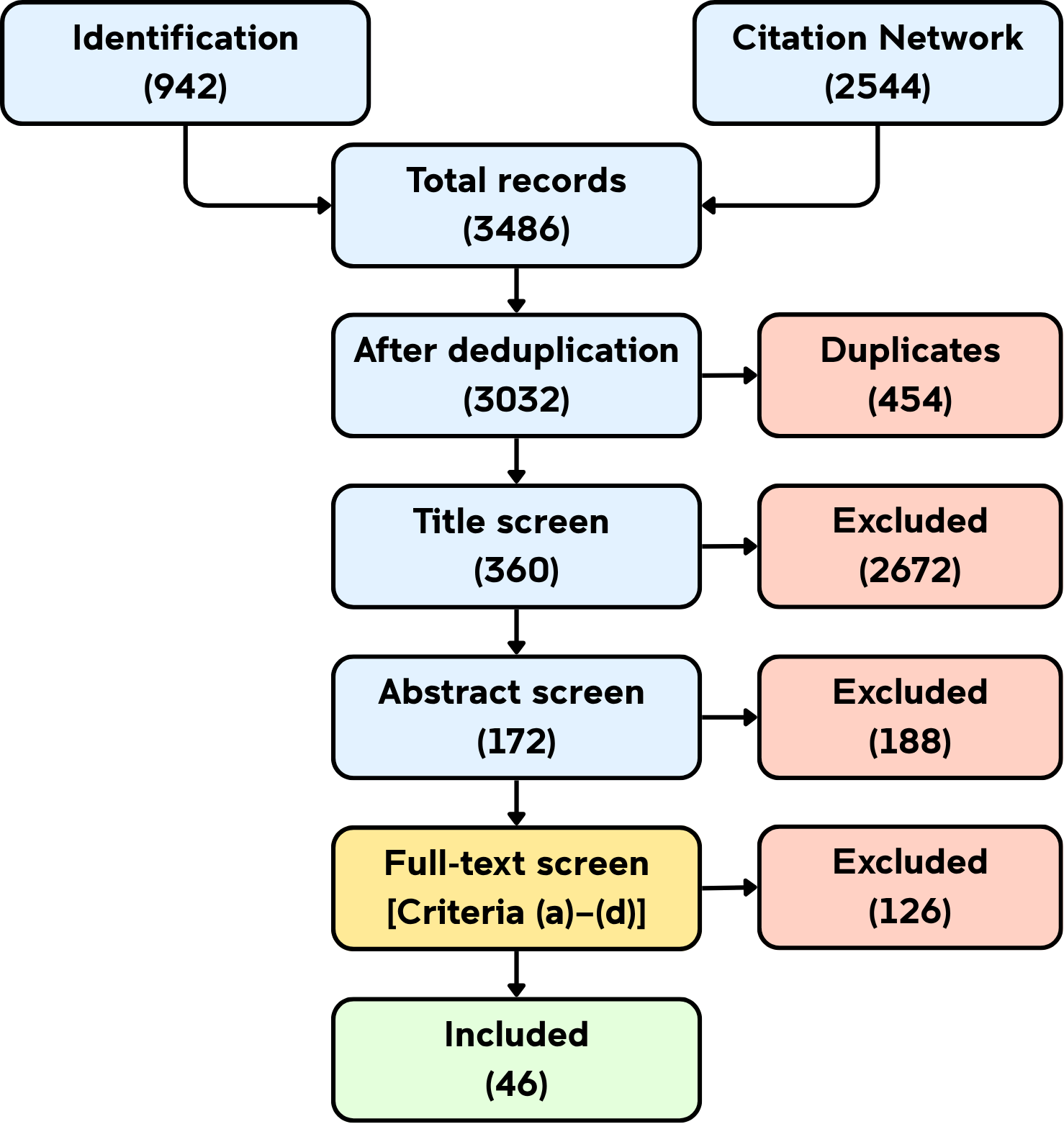}
    \caption{PRISMA-style flow of the search and screening process. Inclusion criteria is described in Section~\ref{sec:search-criteria}.}
    \label{fig:prisma}
\end{figure}

\section{Scope and Search Criteria}
\label{sec:search-criteria}

% To identify NLP research for ED workflows, we conducted a systematic search across major NLP, AI, and selected health venues. Our protocol follows PRISMA-style reporting~\citep{Page2021} with documented search sources, Boolean query terms, sequential screening stages, and inter-annotator agreement on abstract screening. Figure~\ref{fig:prisma} summarises the flow from initial retrieval to final corpus; the subsections below detail each component of the protocol.

\paragraph{Sources.}
We searched the following venues in NLP, AI, and health informatics: the ACL
Anthology\footnote{\url{https://aclanthology.org}}, the AAAI Digital
Library\footnote{\url{https://aaai.org/aaai-publications/}}, the NeurIPS
proceedings\footnote{\url{https://papers.nips.cc}}, the ACM Digital
Library\footnote{\url{https://dl.acm.org/}}, the ISCA
Archive\footnote{\url{https://www.isca-archive.org/}}, the Nature
Portfolio\footnote{\url{https://www.nature.com/}}, and PLOS
ONE\footnote{\url{https://journals.plos.org/plosone/}}.
The initial search was completed in August 2025; relevant papers published in
these venues after this date were retrospectively included up to the submission
date.

\paragraph{Search strategy.}
We queried each source using the terms \textit{emergency department},
\textit{emergency room}, \textit{triage}, \textit{clinical}, \textit{hospital}, \textit{natural language processing},
\textit{NLP}, \textit{artificial intelligence}, \textit{machine learning}, and
\textit{clinical}, together with their morphological variants and domain
synonyms (e.g.,~\textit{ED}, \textit{ER}, \textit{A\&E}).
Terms were combined with \textsc{and}/\textsc{or} operators;
at least one of \textit{emergency department}, \textit{emergency room},
\textit{triage}, \textit{clinical} or \textit{hospital} was required in all queries.
We supplemented keyword search with backward and forward citation snowballing
on all retrieved papers.

\paragraph{Inclusion criteria.}
We follow PRISMA-style reporting~\citep{Page2021} with sequential screening 
stages. A paper was included if and only if it satisfied all of the following:
\begin{enumerate}[label=(\alph*), leftmargin=1.5em, itemsep=0pt, parsep=0pt]
      \item \textit{ED phase alignment.} The task maps to at least one of the three 
      ED phases defined in Appendix~\ref{sec:app-phases}.
      \item \textit{ED-relevant language resource.} The work uses a language 
      resource that could plausibly arise in an ED setting (e.g., triage notes, 
      chief complaints, symptom descriptions, diagnostic impressions, or 
      patient-clinician interactions) covering presenting complaints or diagnoses 
      representative of ED visits~\citep{Raven2013}.
      \item \textit{NLP contribution.} The work proposes or evaluates an NLP 
      method as a primary research contribution, independent of shared-task 
      participation.
      \item \textit{ED-anchored evaluation.} The task or evaluation setting is 
      anchored to ED workflows, evidenced by ED-specific data, ED-specific tasks, 
      or explicit acute-care framing, rather than reliance on general clinical 
      benchmarks or broad inpatient corpora (e.g., MIMIC-III;~\citealt{mimic-iii}) 
      alone.
\end{enumerate}

\paragraph{Screening process.}
Figure~\ref{fig:prisma} summarises the search and screening flow across three sequential stages.
\textit{Stage~1}: title screening to remove clearly off-topic results.
\textit{Stage~2}: abstract screening to assess topical fit.
\textit{Stage~3}: full-text review against the inclusion criteria below.
All three stages were conducted by the lead author.
To assess reliability, a second author (an NLP researcher) independently
screened a random 20\% sample of abstracts reaching Stage~2, yielding a
Cohen's $\kappa$ of 0.80 (rounded to two decimal places), indicating substantial 
agreement.
Disagreements at Stage~2 were resolved by discussion with a third author,
an emergency medicine clinician; Stage~3 decisions were made by the lead
author. The final corpus comprises 46 peer-reviewed papers that explicitly formulate NLP tasks within emergency department workflows.

\section{Papers Included in the Survey}
\label{sec:paper-list}

Tables~\ref{tab:appendix_full_catalog_1} and~\ref{tab:appendix_full_catalog_2} provide a paper-level catalogue of all 46 included papers. We outline the trends across three dimensions: clinical settings, training paradigms, and evaluation paradigms.

\begin{sidewaystable*}[p]
\setlength{\tabcolsep}{3pt}
\begin{adjustbox}{width=0.92\linewidth, center}
\small
\begin{tabular}{llccccccccccccccccccccc}
\toprule
& \textbf{ED}
& \multicolumn{3}{c}{\textbf{Phases}}
& \multicolumn{3}{c}{\textbf{Languages}}
& \multicolumn{3}{c}{\textbf{Input modality}}
& \multicolumn{8}{c}{\textbf{Training paradigm}}
& \multicolumn{4}{c}{\textbf{Evaluation paradigm}} \\
\cmidrule(lr){3-5}
\cmidrule(lr){6-8}
\cmidrule(lr){9-11}
\cmidrule(lr){12-19}
\cmidrule(lr){20-23}
\textbf{Paper} & \textbf{Tag}
& \textbf{Triage} & \textbf{Diag./Assess.} & \textbf{Disposition}
& \textbf{English} & \textbf{Chinese} & \textbf{Others}
& \textbf{Text} & \textbf{Struct.} & \textbf{Image}
& \textbf{Pre.} & \textbf{TL} & \textbf{Sup.} & \textbf{RL} & \textbf{KG} & \textbf{RAG} & \textbf{Prompt} & \textbf{Agent}
& \textbf{Retro.} & \textbf{Prosp.} & \textbf{+Clin.} & \textbf{Deploy.} \\
\midrule

\citet{wei-etal-2018-task}              & \edadjacent &     & \ok &     &     & \ok &     & \ok &     &     &     &     &     & \ok &     &     &     &     &     &     &     &     \\
\citet{jeblee-etal-2019-extracting}     & \edanchored &     & \ok &     & \ok &     &     & \ok &     &     &     &     & \ok &     &     &     &     &     & \ok &     &     &     \\
\citet{Fernandes2020}                   & \ednative &     &     & \ok & \ok &     & \ok &     & \ok &     &     &     & \ok &     &     &     &     &     & \ok &     &     &     \\
\citet{joshi-etal-2020-dr}              & \edadjacent &     & \ok &     & \ok &     &     & \ok &     &     &     &     & \ok &     &     &     &     &     & \ok &     & \ok &     \\
\citet{li-etal-2020-pharmmt}            & \edanchored &     &     & \ok & \ok &     &     & \ok &     &     &     & \ok &     &     &     &     &     &     & \ok &     & \ok &     \\
\citet{lovelace-mortazavi-2020-learning}& \edanchored &     & \ok &     & \ok &     &     & \ok &     & \ok &     & \ok &     &     &     &     &     &     & \ok &     &     &     \\
\citet{rahman2020grace}                 & \ednative &     & \ok &     & \ok &     &     & \ok &     &     &     &     & \ok &     &     &     &     &     &     &     &     &     \\
\citet{pmlr-v126-schloss20a}            & \edanchored &     & \ok &     & \ok &     &     & \ok &     &     &     &     & \ok &     &     &     &     &     & \ok &     &     &     \\
\citet{krishna2021generating}           & \edanchored &     & \ok &     & \ok &     &     & \ok &     &     &     & \ok &     &     &     &     &     &     & \ok &     & \ok &     \\
\citet{lin2021graph}                    & \edadjacent &     & \ok &     &     & \ok &     &     & \ok &     &     &     &     &     & \ok &     &     &     &     &     &     &     \\
\citet{fansi-etal-2022-towards-trustworthy} & \edadjacent &  & \ok &     & \ok &     &     &     & \ok &     &     &     &     & \ok &     &     &     &     &     &     &     &     \\
\citet{grambow-etal-2022-domain}        & \edanchored &     & \ok &     & \ok &     &     & \ok &     &     & \ok &     &     &     &     &     &     &     & \ok &     &     &     \\
\citet{li-etal-2022-diaformer}          & \edadjacent &     & \ok &     &     & \ok &     & \ok & \ok &     &     &     & \ok &     &     &     &     &     &     &     &     &     \\
\citet{liu2022retrieve}                 & \edanchored &     &     & \ok & \ok &     &     & \ok &     &     &     &     &     &     &     &     &     &     & \ok &     & \ok &     \\
\citet{naseem2022incorporating}         & \edadjacent &     & \ok &     & \ok &     &     & \ok & \ok &     &     &     &     &     & \ok &     &     &     &     &     & \ok &     \\
\citet{su2022extract}                   & \edanchored &     & \ok &     & \ok &     &     & \ok &     &     &     & \ok &     &     &     &     &     &     & \ok &     &     &     \\
\citet{chen-etal-2023-clinical}         & \edanchored &     & \ok &     & -   & -   & -   & \ok &     &     &     & \ok &     &     &     &     &     &     &     &     &     &     \\
\citet{Chen2023}                        & \edanchored &     & \ok &     & \ok &     &     & \ok &     &     & \ok &     &     &     &     &     &     &     & \ok &     &     &     \\
\citet{dou2023plugmed}                  & \edadjacent &     & \ok &     & \ok & \ok &     & \ok & \ok &     &     &     &     &     &     &     &     & \ok &     &     & \ok &     \\
\citet{eremeev-etal-2023-injecting}     & \edanchored &     &     & \ok & -   & -   & -   & \ok & \ok &     &     &     &     &     & \ok &     &     &     & \ok &     & \ok &     \\
\citet{hou2023mtdiag}                   & \edadjacent &     & \ok &     &     & \ok &     & \ok & \ok &     &     &     & \ok &     &     &     &     &     &     &     &     &     \\
\citet{nair2023generating}              & \edanchored &     & \ok &     & \ok &     &     & \ok &     &     &     &     &     &     &     &     & \ok &     & \ok &     & \ok &     \\
\citet{singh-etal-2023-large}           & \edanchored &     & \ok &     & \ok &     &     & \ok &     &     &     & \ok &     &     &     &     &     &     & \ok &     &     &     \\
\citet{wang-etal-2023-coad}             & \edadjacent &     & \ok &     &     & \ok &     & \ok & \ok &     &     &     & \ok &     &     &     &     &     &     &     &     &     \\
\citet{xu2023medical}                   & \edadjacent &     & \ok &     &     & \ok &     & \ok &     &     &     &     & \ok &     &     &     &     &     &     &     & \ok &     \\

\bottomrule
\end{tabular}
\end{adjustbox}
\caption{Full catalog of included papers, Part~1 (2018--2023), covering papers from the field's earliest contributions through the emergence of LLM-based approaches. Each paper is annotated across six column groups. The \textbf{Tag} column indicates ED-relatedness: {\ednative} = ED-native (ED data + ED task), {\edanchored} = ED-anchored (non-ED data + ED-relevant task), {\edadjacent} = ED-adjacent (telemedicine/online/synthetic data + transferable task). The \textit{Phases} group indicates which stage of the ED workflow the paper addresses: Triage, Diagnosis and Assessment (Diag./Assess.), or Disposition. The \textit{Languages} and \textit{Input modality} groups record the data language and whether the model consumes free text, structured records, or images; a dash in the language columns means the source language is unspecified. The \textit{Training paradigm} group captures the dominant modelling strategy, from classical supervised fine-tuning (Sup.) and transfer learning (TL) through pretraining (Pre.), reinforcement learning (RL), knowledge-grounded generation (KG), retrieval-augmented generation (RAG), prompt-based or in-context learning (Prompt), and agent-based frameworks (Agent). The \textit{Evaluation paradigm} group encodes study rigour: Retro.\ marks papers that evaluate on real retrospective patient records; Prosp.\ marks the one prospective study in the corpus; papers with neither mark rely on simulation or synthetic data. +Clin.\ marks papers whose evaluation includes clinician or human judgment beyond automated metrics; Deploy.\ marks systems reported as clinically deployed.}
\label{tab:appendix_full_catalog_1}
\end{sidewaystable*}

\begin{sidewaystable*}[p]
\centering
\setlength{\tabcolsep}{3pt}
\begin{adjustbox}{width=0.92\linewidth, center}
\small
\begin{tabular}{llccccccccccccccccccccc}
\toprule
& \textbf{ED}
& \multicolumn{3}{c}{\textbf{Phases}}
& \multicolumn{3}{c}{\textbf{Languages}}
& \multicolumn{3}{c}{\textbf{Input modality}}
& \multicolumn{8}{c}{\textbf{Training paradigm}}
& \multicolumn{4}{c}{\textbf{Evaluation paradigm}} \\
\cmidrule(lr){3-5}
\cmidrule(lr){6-8}
\cmidrule(lr){9-11}
\cmidrule(lr){12-19}
\cmidrule(lr){20-23}
\textbf{Paper} & \textbf{Tag}
& \textbf{Triage} & \textbf{Diag./Assess.} & \textbf{Disposition}
& \textbf{English} & \textbf{Chinese} & \textbf{Others}
& \textbf{Text} & \textbf{Struct.} & \textbf{Image}
& \textbf{Pre.} & \textbf{TL} & \textbf{Sup.} & \textbf{RL} & \textbf{KG} & \textbf{RAG} & \textbf{Prompt} & \textbf{Agent}
& \textbf{Retro.} & \textbf{Prosp.} & \textbf{+Clin.} & \textbf{Deploy.} \\
\midrule

\citet{dou2024integrating}              & \edadjacent &     & \ok &     & \ok & \ok &     & \ok & \ok &     &     &     &     &     &     &     &     & \ok &     &     &     &     \\
\citet{he-etal-2024-bp4er}              & \edadjacent &     & \ok &     &     & \ok &     & \ok &     &     &     &     &     &     &     &     &     & \ok &     &     & \ok &     \\
\citet{Ip2024}                          & \ednative &     &     & \ok & \ok &     & \ok &     &     & \ok &     &     & \ok &     &     &     &     &     &     & \ok &     &     \\
\citet{lu-etal-2024-triageagent}        & \ednative & \ok &     &     & \ok &     &     & \ok &     &     &     &     &     &     &     &     &     & \ok &     &     &     &     \\
\citet{maschhur-etal-2024-towards}      & \ednative & \ok &     &     &     &     & \ok &     & \ok &     &     & \ok &     &     &     &     &     &     & \ok &     &     &     \\
\citet{sun-etal-2024-edcopilot}         & \ednative &     &     & \ok & \ok &     &     & \ok & \ok &     &     &     &     & \ok &     &     &     & \ok & \ok &     &     &     \\
\citet{xu2024reasoning}                 & \edadjacent &     & \ok &     &     & \ok &     & \ok &     &     &     &     &     &     & \ok &     &     &     &     &     &     &     \\
\citet{yuan-etal-2024-continued}        & \edanchored &     & \ok &     & \ok &     &     & \ok &     &     & \ok &     &     &     &     &     &     &     & \ok &     &     &     \\
\citet{Chen2025}                        & \edadjacent &     & \ok &     & \ok &     &     & \ok & \ok &     &     &     &     &     &     &     &     & \ok &     &     &     &     \\
\citet{Gaber2025}                       & \ednative & \ok &     &     & \ok &     &     & \ok & \ok &     &     &     &     &     &     & \ok &     &     & \ok &     &     &     \\
\citet{gatto-etal-2025-follow}          & \edanchored &     & \ok &     & \ok &     &     & \ok & \ok &     &     &     & \ok &     &     &     &     &     & \ok &     &     &     \\
\citet{heal-www-2025}                   & \ednative & \ok &     &     & \ok &     &     & \ok & \ok &     &     &     &     &     &     &     &     & \ok &     &     &     &     \\
\citet{jia2025medikal}                  & \edadjacent &     & \ok &     &     & \ok &     & \ok & \ok &     &     &     &     &     & \ok &     &     &     & \ok &     &     &     \\
\citet{ju-lee-2025-prediction}          & \edadjacent &     & \ok &     &     & \ok &     & \ok & \ok &     &     &     &     &     &     &     &     & \ok &     &     &     &     \\
\citet{Liu2025}                         & \ednative & \ok &     &     & \ok &     &     & \ok & \ok &     &     & \ok &     &     &     &     &     &     & \ok &     &     &     \\
\citet{nicolson-etal-2025-impact}       & \ednative &     & \ok &     & \ok &     &     & \ok & \ok &     &     &     & \ok &     &     &     &     &     & \ok &     &     &     \\
\citet{qin-etal-2025-listening}         & \edadjacent &     & \ok &     & \ok &     &     & \ok &     &     &     &     & \ok &     &     &     &     &     &     &     &     &     \\
\citet{rose-etal-2025-meddxagent}       & \edadjacent &     & \ok &     & \ok &     &     & \ok & \ok &     &     &     &     &     &     &     &     & \ok &     &     &     &     \\
\citet{sun-etal-2025-enhancing-medical} & \edadjacent &     & \ok &     & \ok & \ok &     & \ok & \ok &     &     &     &     &     &     &     &     & \ok &     &     &     &     \\
\citet{wang-etal-2025-towards-adapting} & \edanchored &     & \ok &     & \ok &     &     & \ok &     &     & \ok &     &     &     &     &     &     &     & \ok &     & \ok &     \\
\citet{xia2025mmedrag}                  & \edanchored &     & \ok &     & \ok &     &     & \ok &     & \ok &     &     &     &     &     & \ok &     &     & \ok &     &     &     \\

\bottomrule
\end{tabular}
\end{adjustbox}
\caption{Full catalog of included papers, Part~2 (2024--2025). Column groups and coding conventions follow Table~\ref{tab:appendix_full_catalog_1}.}
\label{tab:appendix_full_catalog_2}
\end{sidewaystable*}
\section{Reported Performances}
\label{app:performance}
\newcolumntype{L}[1]{>{\raggedright\arraybackslash}p{#1}}

\begin{table*}[t!]
\centering
\small
\setlength{\tabcolsep}{5pt}
\renewcommand{\arraystretch}{1.15}
\begin{tabular}{@{}L{2.4cm} L{2.0cm} L{3.2cm} L{5.4cm}@{}}
\toprule
\textbf{Task} & \textbf{Training Paradigm} & \textbf{Paper} & \textbf{Reported performance} \\
\midrule

\multirow{5}{*}{Triage classification}
& \multirow{2}{*}{Transfer learning}
    & \citet{maschhur-etal-2024-towards} & F1: 0.63 \\
&   & \citet{Liu2025} & AUC: 0.88 \\
\cmidrule(lr){2-4}
& \multirow{2}{*}{Agent-based}
    & \citet{lu-etal-2024-triageagent} & Discordance: 0.19; Acc.: 0.81 \\
&   & \citet{heal-www-2025} & MAE: 0.27; Acc.: 0.73 \\
\cmidrule(lr){2-4}
& Retrieval-augmented
    & \citet{Gaber2025} & Acc.: 0.66 \\

\midrule

\multirow{13}{*}{\parbox{2.4cm}{Clinical interaction summarisation}}
& \multirow{4}{*}{\parbox{2.0cm}{Neural architectures}}
    & \citet{jeblee-etal-2019-extracting} & F1: 0.60 \\
&   & \citet{pmlr-v126-schloss20a} & F1: 0.44; AUROC: 0.83 \\
&   & \citet{joshi-etal-2020-dr} & R-L: 0.55 \\
&   & \citet{rahman2020grace} & F1: 0.63 \\
\cmidrule(lr){2-4}
& \multirow{4}{*}{Transfer learning}
    & \citet{krishna2021generating} & R-L: 0.38 \\
&   & \citet{su2022extract} & R-L: 0.34 \\
&   & \citet{chen-etal-2023-clinical}\textsuperscript{a} & Acc.\ gain: $+10.7$ \\
&   & \citet{singh-etal-2023-large} & R-L: 0.64; Fact-C: 0.69 \\
\cmidrule(lr){2-4}
& \multirow{4}{*}{Pretraining}
    & \citet{grambow-etal-2022-domain} & R-L: 0.27; UMLS-F1: 0.39 \\
&   & \citet{Chen2023} & R-L: 0.27; UMLS-F1: 0.37 \\
&   & \citet{yuan-etal-2024-continued}\textsuperscript{b} & Missed: 4.3; Incorrect: 0.85; Irrelevant: 0.30 \\
&   & \citet{wang-etal-2025-towards-adapting} & R-L: 0.36; Completeness: 4.3 \\
\cmidrule(lr){2-4}
& Prompt-based
    & \citet{nair2023generating}\textsuperscript{c} & GPT-F1: 0.62 \\

\bottomrule
\end{tabular}
\caption{Reported performance across the reviewed clinical tasks (Part 1 of 3): triage classification and clinical interaction summarisation. See Table~\ref{tab:perf-part3} for footnotes and abbreviations.}
\label{tab:perf-part1}
\end{table*}

\begin{table*}[t!]
\centering
\small
\setlength{\tabcolsep}{5pt}
\renewcommand{\arraystretch}{1.15}
\begin{tabular}{@{}L{2.4cm} L{2.0cm} L{3.2cm} L{5.4cm}@{}}
\toprule
\textbf{Task} & \textbf{Training Paradigm} & \textbf{Paper} & \textbf{Reported performance} \\
\midrule

\multirow{19}{*}{\parbox{2.4cm}{Automatic diagnosis}}
& \multirow{2}{*}{\parbox{2.0cm}{Reinforcement learning}}
    & \citet{wei-etal-2018-task} & Acc.: 0.65; Turns: 5.11 \\
&   & \citet{fansi-etal-2022-towards-trustworthy} & Acc.: 99.2; Turns: 5.47 \\
\cmidrule(lr){2-4}
& \multirow{6}{*}{Supervised}
    & \citet{li-etal-2022-diaformer} & Acc.: 0.77; Turns: 14.3 \\
&   & \citet{wang-etal-2023-coad} & Acc.: 0.69; Turns: 13.25 \\
&   & \citet{hou2023mtdiag} & Acc.: 81.4; Recall: 87.6; Turns: 14.8 \\
&   & \citet{xu2023medical} & R-1: 0.30; B-1: 0.43; Entity-F1: 0.23 \\
&   & \citet{gatto-etal-2025-follow}\textsuperscript{d} & Questions: 36 \\
&   & \citet{qin-etal-2025-listening} & R-1: 0.28; B-1: 0.44; Entity-F1: 0.25 \\
\cmidrule(lr){2-4}
& \multirow{4}{*}{\parbox{2.0cm}{Knowledge-grounded}}
    & \citet{naseem2022incorporating} & B-2: 0.15; Human eval.: 4.0 \\
&   & \citet{lin2021graph}\textsuperscript{e} & BLEU avg.: 0.36; Entity-F1: 0.47 \\
&   & \citet{jia2025medikal} & F1: 0.37 \\
&   & \citet{xu2024reasoning} & B-4: 0.21; R-2: 0.13; Entity-F1: 0.24 \\
\cmidrule(lr){2-4}
& \multirow{7}{*}{Agent-based}
    & \citet{dou2024integrating}\textsuperscript{f} & Symptoms: 0.25; Tests: 0.42; Dx: 0.55 \\
&   & \citet{Chen2025} & Acc.: 0.34 \\
&   & \citet{rose-etal-2025-meddxagent}\textsuperscript{g} & GTPA@1: 0.72; Avg.\ rank: 2.2 \\
&   & \citet{ju-lee-2025-prediction} & Recall@1: 0.75; Recall@3: 0.97 \\
&   & \citet{dou2023plugmed} & R-L: 0.16; BERTScore: 0.61 \\
&   & \citet{he-etal-2024-bp4er} & B-4: 0.23; R-2: 0.22 \\
&   & \citet{sun-etal-2025-enhancing-medical} & B-4: 0.22; R-2: 0.13; Entity-F1: 0.22 \\

\bottomrule
\end{tabular}
\caption{Reported performance across the reviewed clinical tasks (Part 2 of 3): automatic diagnosis. See Table~\ref{tab:perf-part3} for footnotes and abbreviations.}
\label{tab:perf-part2}
\end{table*}

\begin{table*}[t!]
\centering
\small
\setlength{\tabcolsep}{5pt}
\renewcommand{\arraystretch}{1.15}
\begin{tabular}{@{}L{2.4cm} L{2.0cm} L{3.2cm} L{5.4cm}@{}}
\toprule
\textbf{Task} & \textbf{Training Paradigm} & \textbf{Paper} & \textbf{Reported performance} \\
\midrule

\multirow{3}{*}{\parbox{2.4cm}{Radiology report generation}}
& Transfer learning
    & \citet{lovelace-mortazavi-2020-learning} & B-4: 0.15; CIDEr: 0.31; R-L: 0.32 \\
\cmidrule(lr){2-4}
& Supervised
    & \citet{nicolson-etal-2025-impact} & R-L: 0.26; B-4: 0.05; BERTScore: 0.25 \\
\cmidrule(lr){2-4}
& \multirow{1}{*}{\parbox{2.0cm}{Retrieval-augmented}}
    & \citet{xia2025mmedrag}\textsuperscript{h} & R-L: 0.19; BLEU avg.: 0.23; METEOR: 0.27 \\

\midrule

\multirow{3}{*}{\parbox{2.4cm}{Disposition prediction}}
& \multirow{2}{*}{Supervised}
    & \citet{Fernandes2020} & Recall: 0.82; AUROC: 0.91; AUPRC: 0.30 \\
&   & \citet{Ip2024} & AUROC: 0.71; AUPRC: 0.64 \\
\cmidrule(lr){2-4}
& Agent-based
    & \citet{sun-etal-2024-edcopilot} & F1: 0.32; AUC: 0.77 \\

\midrule

\multirow{2}{*}{\parbox{2.4cm}{Discharge instruction generation}}
& \multirow{1}{*}{\parbox{2.0cm}{Knowledge-grounded}}
    & \citet{eremeev-etal-2023-injecting} & BERTScore: 0.32; Concept-F1: 0.76; PPL: 6.96 \\
\cmidrule(lr){2-4}
& Transfer learning
    & \citet{li-etal-2020-pharmmt} & B-4: 0.60; METEOR: 0.76 \\

\bottomrule
\multicolumn{4}{@{}p{13.4cm}@{}}{\footnotesize
\textbf{Notes.} Values are in the original scale used by each paper and are not directly comparable across rows because datasets, label spaces, task formulations, and evaluation protocols differ. Where a paper reported results on multiple datasets, the table reports the average value or the primary result emphasised by the paper.
\textsuperscript{a} Reports improvement over a baseline rather than an absolute score.
\textsuperscript{b} Reports the average number of missed, incorrect, and irrelevant facts.
\textsuperscript{c} Evaluation uses GPT-based scoring.
\textsuperscript{d} Reports the average number of follow-up questions.
\textsuperscript{e} BLEU avg.\ denotes the average of BLEU-1/2/3/4.
\textsuperscript{f} Reports the probability of selecting the correct symptom, test, or diagnosis category.
\textsuperscript{g} GTPA@1 denotes the top-1 Ground-Truth Path Accuracy.
\textsuperscript{h} BLEU avg.\ denotes the average of BLEU-1/2/4.
Abbreviations: Acc.\ = accuracy; AUC/AUROC = area under the ROC curve; AUPRC = area under the precision-recall curve; MAE = mean absolute error; R-1/R-2/R-L = ROUGE-1/2/L; B-1/B-2/B-4 = BLEU-1/2/4; CIDEr = Consensus-based Image Description Evaluation; PPL = perplexity.
} \\
\end{tabular}
\caption{Reported performance across the reviewed clinical tasks (Part 3 of 3): radiology report generation, disposition prediction, and discharge instruction generation.}
\label{tab:perf-part3}
\end{table*}

Tables~\ref{tab:perf-part1}--\ref{tab:perf-part3} summarise reported performance across the clinical tasks reviewed in this survey. These values should not be interpreted as directly comparable because studies differ substantially in datasets, task formulations, label spaces, and evaluation protocols.

Three broad patterns emerge. First, performance is strongly task dependent: constrained prediction tasks such as triage and disposition prediction generally report stronger headline scores than open-ended generation tasks such as clinical interaction summarisation and radiology report generation. Second, automatic diagnosis shows the greatest evaluation heterogeneity, reflecting the breadth of formulations used in prior work, from disease prediction and symptom acquisition to dialogue generation and multi-step reasoning. Third, newer paradigms such as agent-based, retrieval-augmented, and prompt-based methods are increasingly visible, but their gains remain difficult to interpret consistently because evaluation settings vary widely.

The distribution of results also suggests a clear pattern in training paradigms. Supervised and transfer learning methods remain the most established approaches across tasks, particularly for classification and structured generation settings where labels and evaluation criteria are relatively stable. By contrast, agent-based, prompt-based, and retrieval-augmented methods are more common in tasks that require multi-step reasoning, external knowledge, or flexible generation, such as automatic diagnosis and clinical interaction summarisation. Their growing presence points to methodological expansion rather than convergence, while reinforcement learning remains more specialised and appears mainly in diagnosis settings involving sequential decision-making.

Overall, the table highlights the fragmented state of evaluation in ED-focused NLP. Classification tasks are usually assessed with metrics such as accuracy, AUC, AUROC, and AUPRC, whereas summarisation and generation tasks rely more on ROUGE, BLEU, factuality, and concept-level measures. The table should therefore be read as a compact reference for the range of reported outcomes in the literature rather than as evidence of a single best-performing modelling strategy.
\end{document}